%% file: geom_dl_paper.tex
\documentclass{article}
\usepackage{iclr2015}
\usepackage{mathptmx}
\usepackage{graphicx}
\usepackage{amsmath,amsthm,amsfonts, amssymb, dsfont, subfigure}

\newcommand{\pone}{\textbf{P1}}
\newcommand{\ptwo}{\textbf{P2}}

\iclrfinalcopy

\title{A Group Theoretic Perspective on Unsupervised Deep Learning\thanks{This
    research supported in part by the NSF under grant BIGDATA-1251049}}
\author{Arnab Paul \\ 
Intel Corporation \\
\texttt{arnab.paul@intel.com}
\And 
Suresh Venkatasubramanian \\ 
School of Computing, University of Utah\\
\texttt{suresh@cs.utah.edu}
}

\begin{document}
\maketitle

\input geom_dl_paper-extendedabstract
\bibliography{geom_dl_paper}
\bibliographystyle{iclr2015}

\end{document}

%% file: geom_dl_paper-extendedabstract.tex
\section*{Extended Abstract}
The modern incarnation of neural networks,  now popularly known as Deep Learning (DL),
accomplished record-breaking success in processing diverse kinds of signals -
vision, audio, and text. 
In parallel, strong interest has ensued towards constructing a \emph{theory} of DL. 
This paper opens up a group theory based approach, towards a theoretical 
understanding of DL, in particular the unsupervised variant. First we establish how a single layer of unsupervised pre-training can be explained in the light of orbit-stabilizer principle, and then we sketch how the same principle can be extended for multiple layers.

We focus on two key principles that (amongst others) influenced the modern DL resurgence.

\begin{itemize}
\item[(\pone)]  
Geoff Hinton summed this up as follows. 
``In order to do computer vision, first learn how to do computer
graphics". \cite{hinton2007recognize}. In other words, if a network learns a good generative model of its
training set, then it could use the same model for classification.

\item[(\ptwo)]  
Instead of learning an entire network all at once,  learn it one layer at a
time.
\end{itemize}

\begin{figure}
\subfigure[General auto-encoder schematic]{\label{fig:aepic1}\includegraphics[width=1.2in, height=1in]{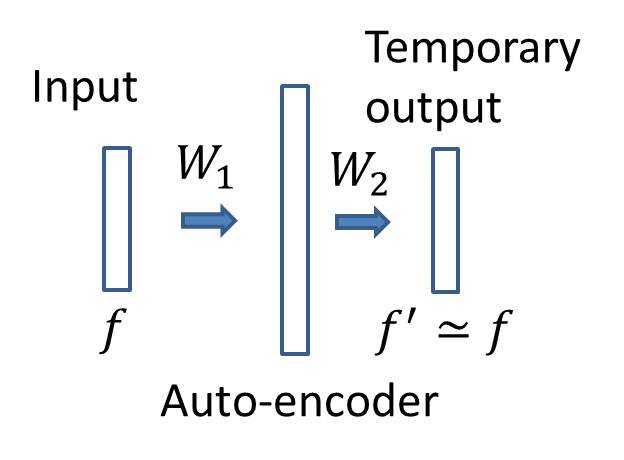}}
\hspace{0.2in}
\subfigure[post-learning behavior of an auto-encoder]{\label{fig:aepic2}\includegraphics[width=1.2in, height=1in]{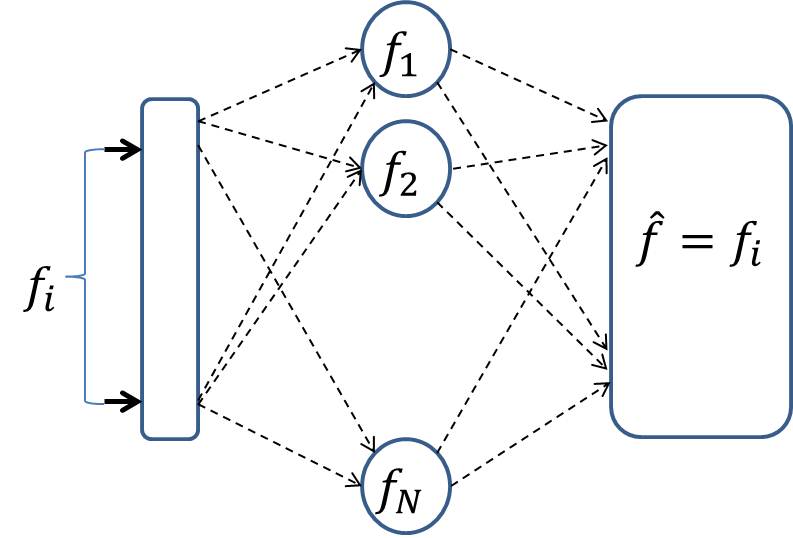}}
\hspace{0.2in}
\subfigure[Alternate Decomposition of a Signal]{\label{fig:decompose}\includegraphics[width=2in,height=1.1in]{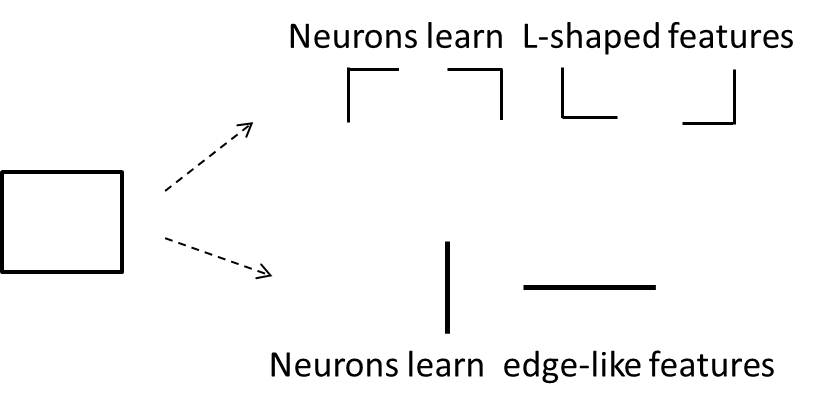}}
\caption{(a) $W_1$ is preserved, $W_2$ discarded \;\;\; (b) Post-learning, each feature is stabilized \;\;\;(c)Alternate ways of  decomposing a signal into simpler features. The neurons could potentially 
learn features in the top row, or the bottom row. Almost surely, the {\em simpler} ones (bottom row) are learned.}
\end{figure}

In each round, the training layer is connected to a temporary output layer and
trained to learn the weights needed to reproduce its input (i.e to solve \pone).
This step -- executed layer-wise, starting with the first hidden layer and
sequentially moving deeper -- is often referred to as pre-training (see
\cite{Hinton06,hinton2007recognize, salakhutdinov2009deep, bengio--dlbook}) and
the resulting layer is called an \emph{autoencoder}.
Figure \ref{fig:aepic1} shows a schematic autoencoder. Its weight set $W_1$ is
learnt by the network. Subsequently when presented with an input $f$, the
network will produce an output $f' \approx f$. At this point the output units
as well as the weight set $W_2$ are discarded.

There is an alternate characterization of \pone.
An autoencoder unit, such as the above, maps an input space to itself.
Moreover, after learning, it is by definition, a \emph{stabilizer}\footnote{ A transformation $T$ is called a stabilizer of an input $f$, if $f' = T(f) = f$.} of the input $f$. 
Now, input signals are often decomposable  into features,
and an autoencoder attempts to find a succinct set of features that all inputs can be decomposed into.
Satisfying \pone  means that the learned configurations can reproduce these features.  
Figure \ref{fig:aepic2} illustrates this post-training behavior. If the hidden units learned features $f_1, f_2, \ldots$, and one of then, say $f_i$, comes back as input, the output must be $f_i$. 
In other words \emph{learning a feature is equivalent to searching for a transformation that stabilizes it}. 

The idea of stabilizers invites an analogy reminiscent of the orbit-stabilizer
relationship studied in the theory of group actions. Suppose $G$ is a group that
acts on a set $X$ by moving its points around (e.g groups of $2 \times 2$
invertible matrices acting over the Euclidean plane). Consider $x \in X$, and
let $O_x$ be the set of all points reachable from $x$ via the group
action. $O_x$ is called an orbit\footnote{The orbit $O_x$ of
  an element $x \in X$ under the action of a group $G$, is defined as the set
  $O_x = \{g(x) \in X | g \in G\}$.}.  A subset of the  group elements may leave $x$
unchanged. This subset $S_x$ (which is also a subgroup), is the stabilizer of
$x$.  If it is possible to define a notion of volume for a group, then there is an inverse relationship between the volumes of
$S_x$ and $O_x$, which holds even if $x$ is actually a subset (as opposed to
being a point). For example, for finite groups, the product of $|O_x|$ and
$|S_x|$ is the order of the group.

The \emph{inverse} relationship between the volumes of orbits and stabilizers
takes on a central role as we connect this back to DL.  
There are many possible ways to decompose signals into smaller features. 
Figure \ref{fig:decompose} illustrates this point: a  rectangle can be
decomposed into L-shaped features or straight-line edges. 

All experiments to date suggest that a neural network is likely to learn the
edges. But why? To answer this, imagine that the space of the autoencoders
(viewed as transformations of the input) form a group.  A batch of learning iterations stops \emph{whenever} a stabilizer is found. 
Roughly speaking, if the search is a
Markov chain (or a guided chain such as MCMC), then 
the bigger a stabilizer, the earlier it will be hit.
The group structure implies that this big stabilizer corresponds to a small orbit. 
Now intuition suggests that the simpler a feature, the smaller is its orbit. 
For example, a line-segment generates many fewer possible shapes
under linear 
deformations than a flower-like shape. An autoencoder then should
learn these \emph{simpler} features first, which falls in line with most experiments (see \cite{lee2009convolutional}).

The intuition naturally extends to a many-layer scenario.
Each  hidden layer finding a feature with a big stabilizer.
But beyond the first level, the inputs no longer inhabit the same space as the
training samples. A ``simple" feature over this new space actually corresponds
to a more complex shape in the space of input samples. This process repeats as
the number of layers increases. In effect, each layer learns ``edge-like
features'' with respect to the previous layer, and from these locally simple
representations we obtain the learned higher-order representation.